\documentclass[final]{cvpr}

\usepackage{times}
\usepackage{epsfig}
\usepackage{graphicx}
\usepackage{amsmath}
\usepackage{amssymb}

\usepackage{amsthm}
\usepackage{subfig}

\usepackage{subfig}
\usepackage{multirow}
\usepackage[ruled,vlined,linesnumbered]{algorithm2e}

\theoremstyle{lemma}

\theoremstyle{prop}

\usepackage{pifont}
\newcommand{\cmark}{\ding{51}}%
\newcommand{\xmark}{\ding{55}}%

\usepackage[pagebackref=true,breaklinks=true,letterpaper=true,colorlinks,bookmarks=false]{hyperref}

\def\confYear{CVPR 2021}

\begin{document}

\title{Basket-based Softmax}

\author{
  Qiang Meng$^1$, Xinqian Gu$^2$, Xiaqing Xu$^1$, Feng Zhou$^1$ \\ 
  $^1$Algorithm Research, Aibee Inc. $^2$ University of Chinese Academy of Sciences \\
}


\maketitle

\begin{abstract}
  Softmax-based losses
  have achieved state-of-the-art performances on various tasks such as face recognition and re-identification. However, these methods highly relied on clean datasets with global labels, which limits their usage in many real-world applications. An important reason is that merging and organizing datasets from various temporal and spatial scenarios is usually not realistic, as noisy labels can be introduced and exponential-increasing resources are required. To address this issue, we propose a novel mining-during-training strategy called Basket-based Softmax (BBS) as well as its parallel version to effectively train models on multiple datasets in an end-to-end fashion. Specifically, for each training sample, we simultaneously adopt similarity scores as the clue to mining negative classes from other datasets, and dynamically add them to assist the learning of discriminative features. Experimentally, we demonstrate the efficiency and superiority of the BBS on the tasks of face recognition and re-identification, with both simulated and real-world datasets.
\end{abstract}

\section{Introduction}

\begin{figure}[htb!]
  \centering
  \includegraphics[clip, trim=0cm 0cm 0cm 0cm, width=0.45\textwidth]{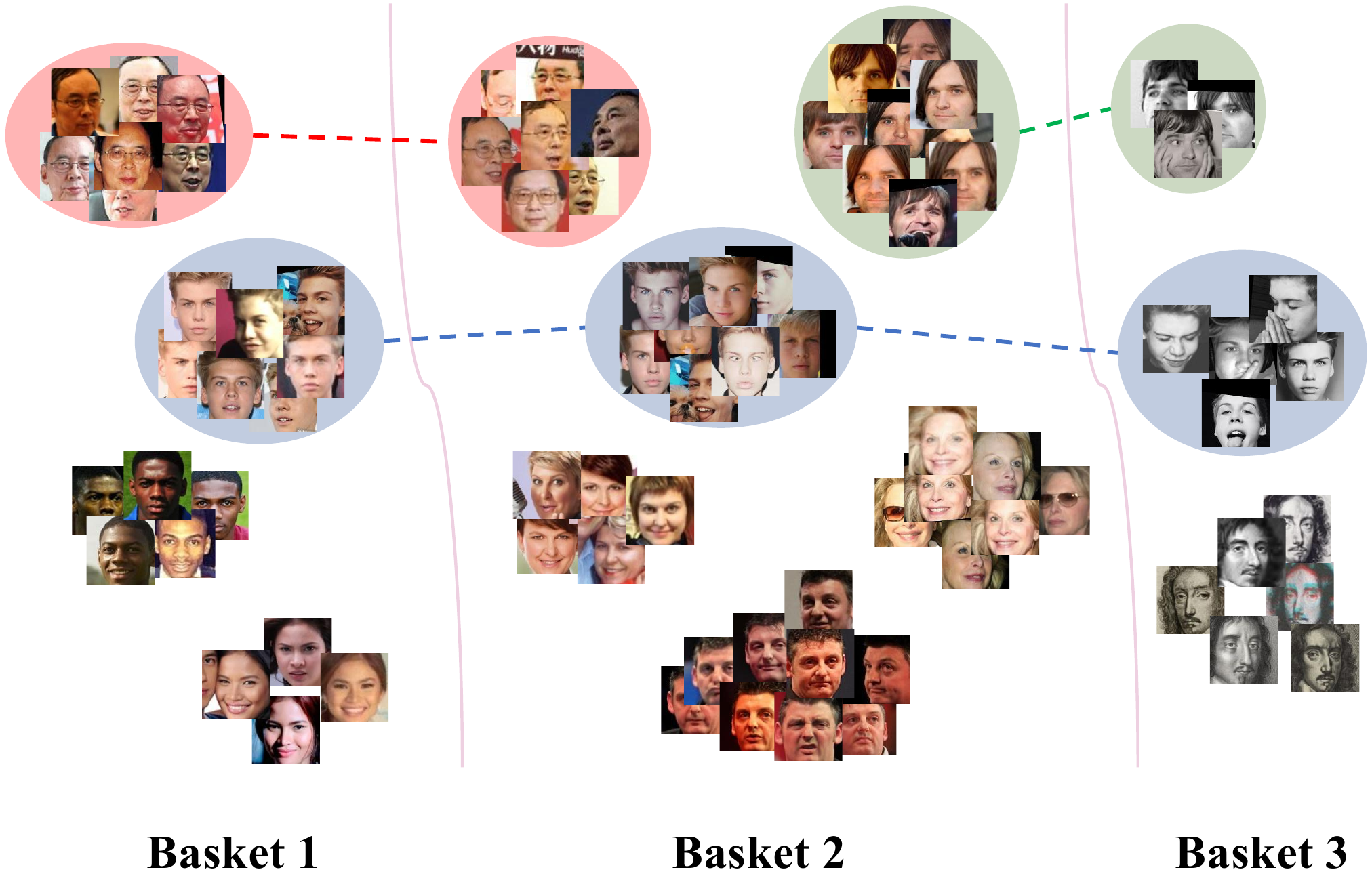}
  \caption{An illustration of the \textbf{M}ultiple \textbf{B}askets with \textbf{C}lass \textbf{O}verlaps (MBCO) problem (simulated by faces from MS1MV2~\cite{Deng2018}). One basket means a well-organized dataset while these baskets can be collected from different temporal/spatial scenarios or capturing conditions (\textit{e.g.}, images in basket 1,2 are captured with RGB cameras while basket 3 are from black-and-white cameras). 
    Overlapped classes between baskets are indicated by dash lines.
  }
  \label{fig:intro}
\end{figure}

Softmax-based losses have achieved state-of-the-art performances on various tasks such as face recognition and re-identification. These methods can be classified into Euclidean margin based loss ~\cite{Liu2016, Ranjan2017, Wang2017} 
 and  cosine margin (angular margin) based loss~\cite{Liu2017, Wang2018, Deng2018, meng2021magface}.
  Compared to pairwised metric learning methods such as triplet loss~\cite{schroff2015facenet} and N-pair~\cite{sohn2016improved} loss, softmax-based losses have the following advantages:
(a). With all negative classes in sight, models trained with softmax-based losses converge more quickly in most cases.
(b). Not affected by sampling strategies and therefore have more stable training processes.
(c). More discriminative features can be learned, and state-of-art performances have been witnessed on various recognition benchmarks.
These characteristics are of great value in many real-world applications where training databases are of large-scale and high efficiencies are needed.

Despite the great success, softmax-based losses suffer from large GPU memory consumption as well as demanding clean datasets with global labels.
Thanks to the development of parallel acceleration, memory assumptions can be highly decreased and that enables training on million level classes on a single machine~\cite{Deng2018}.
However, a clean dataset with global labels is not always available. Instead, in many real-world applications, we may have multiple datasets captured from various temporal/spatial scenarios. Directly merging these datasets, especially for large-scale datasets, is not realistic because noisy labels can be introduced, and exponential-increasing resources are required.
In this case, we call a training dataset as a basket and define the problem as \textbf{M}ultiple \textbf{B}askets with \textbf{C}lass \textbf{O}verlaps (MBCO), where one class is unique in a basket but can appear multiple times across baskets, as illustrated in Fig.~\ref{fig:intro}.
A practically significant but usually ignored question arises: how to simultaneously train softmax losses on multiple baskets with overlapped classes?
We note that the raised problem is different from a well-studied problem~\cite{wu2018light, zhong2019unequal, hu2019noise, wang2019co, deng2020sub} called noisy label problem, which focuses on mislabeled samples inside a class but ignore the noise of one class being assigned with different labels.
To emphasize the importance of this problem, we list a few real-world examples below:
\begin{itemize}
\item In surveillance scenarios
 , merging faces in videos captured within a small-time window is relatively easy as cues such as tracking or re-id results can be relied on. However, merging faces across days
is almost impossible because of the vast volumes of the data as well as the lack of other cues.
\item Merging datasets is extremely difficult when large variations exhibit. For example, face images are highly affected by image acquisition conditions (\eg, illumination, background, blurriness, and low resolution) and factors of the face (\eg, pose, occlusion and expression), and therefore degrades the qualities of human-assessed and similarity-based labels.
\item In some situations, there lack of a clear labelling rules. For instance, whether grouping images from a same identity but with different clothes is still indistinct for person re-identification task.
\item For scenarios with privacy concerns (\eg, collecting faces without  appropriate consents can breach the law), pipelines which can automatically collect data, train and deploy models are preferred. Automatically generating small and clean datasets is relatively easy.
  However, noise will be accumulated in absence of manual interventions with the increasing of data volumes.
  Consequently, models trained on such noisy datasets will encounter performance degradations.
  In this case, models are better trained on multiple clean baskets, instead of a large but noisy dataset.
\end{itemize}

To enable training on multiple baskets with softmax-based losses, we propose a framework called Basket-based Softmax (BBS) which works in an end-to-end fashion and can be applied to most softmax-based losses. Specifically, we simultaneously use the similarity scores defined in softmax losses as the clue to mining negative classes from other baskets, and dynamically add them to guide the learning of more discriminative features.  To better fit the large-scale issue in practical applications, we further modified the BBS to the parallel version. 
Extensive experiments are conducted to verify the superiority and efficiency of our proposed method.
We summarize our contributions as follows:
\begin{enumerate}
\item We raise the importance of the Multiple Baskets with Class Overlaps (MBCO) problem which is usually ignored by the academic community but can occur frequently in many real-world applications.
  Datasets collected from various temporal and spatial scenarios can be of tremendous size and labels are locally assigned in each basket.
  Directly assigning global classes can introduce noisy labels and requires exponential-increasing resources.

\item An efficient and end-to-end mining-during-training framework called BBS is proposed, where similarity scores are adopted as the cue to dynamically mining negative classes across baskets. Our proposed BBS can be applied to the majority softmax-based losses and trained on multiple baskets simultaneously. Besides that, we also introduce the parallel version to enables the training of million level classes.

\item Experimentally, we modified the common-used Softmax, CosFace~\cite{Wang2018} and ArcFace~\cite{Deng2018} to our BBS framework. Extensive experiments are conducted on face recognition and person re-identification and verify the superiority of BBS, with both simulated and real-world training datasets.
\end{enumerate}

\section{Related Works}

\setlength{\tabcolsep}{4pt}
\begin{table*}[htb!]
  \begin{center}
    \footnotesize{
      \begin{tabular}{c|ccccccc}
        \hline
        Method & Softmax  & L-Softmax~\cite{Liu2016} &$l_2$-softmax~\cite{Ranjan2017} & NormFace~\cite{Wang2017} & SphereFace~\cite{Liu2017} &  CosFace~\cite{Wang2018}&   ArcFace~\cite{Deng2018} \\
        \hline
        bias       & \cmark & \xmark & \cmark & \xmark & \xmark &  \xmark & \xmark \\
        $s$        & \xmark & \xmark & \cmark & \cmark & \xmark &  \cmark & \cmark \\
        $f(W,x)$ & $W^Tx$  & $\|W\|\|x\|\psi(\theta)$ & $\|W\|\cos(\theta)$ & $\cos(\theta)$ & $\|x\|\psi(\theta)$ & $cos(\theta)-m$ & $ \cos(\theta+m)$\\
        $g(W,x)$ & $W^Tx$ & $\|W\|\|x\|\cos(\theta)$ & $\|W\|\cos(\theta)$ & $\cos(\theta)$ & $\|x\|\cos(\theta)$ & $\cos(\theta)$ &  $\cos(\theta)$\\
        \hline
      \end{tabular}
    }
  \end{center}
  \caption{A unified perspective of softmax-based losses. Here $\theta = \frac{W^Tx}{\|W\|\|x\|}$ and the \cmark/\xmark  \ denotes used or not in the corresponding work. For L-softmax and SphereFace, the function $\psi$ is defined as $\psi(\theta) = (-1)^k \cos(m\theta)-2k, \theta \in [\frac{k\pi}{m}, \frac{(k+1)\pi}{m}], k\in [0,m-1], m\geq 1$.}  \label{table:method}
\end{table*}
\setlength{\tabcolsep}{1.4pt}

\subsection{Deep Face Recognition}

Recent years have witnessed the breakthrough of deep convolutional face recognition techniques~\cite{Liu2017, Wang2018, Deng2018, xu2021searching, meng2021poseface, meng2021magface, meng2021lce}.
Most of early works rely on metric-learning based loss, including contrastive loss~\cite{chopra2005learning}, triplet loss~\cite{schroff2015facenet}, n-pair loss~\cite{sohn2016improved}, angular loss~\cite{wang2017deep}, \etc.
Suffering from the combinatorial explosion in the number of face triplets, embedding-based method is usually inefficient in training on large-scale dataset.
Therefore, the main body of research in deep face recognition has focused on devising more efficient and effective softmax-based loss.
Wen \etal~\cite{wen2016discriminative} develop a center loss to learn centers for each identity to enhance the intra-class compactness.
$L_2$-softmax~\cite{Ranjan2017} and NormFace~\cite{Wang2017} study the necessity of the normalization operation and applied $L_2$ normalization constraint on both features and weights.


From then on, several angular margin-based losses and progressively improve the performance on various benchmarks to the newer level. SphereFace~\cite{Liu2017} introduced angular margin to softmax loss and achieved discriminative features. To overcome the optimization difficulty of SphereFace, CosFace~\cite{Wang2018} moves the angular margin into cosine space. In ArcFace~\cite{Deng2018},  decision boundary is directly maximized in angular (arc) space based on the normalized weights and features, and they achieve state-of-the-art performances on current benchmarks.

\subsection{Person Re-identification}

Person re-identification (re-id) aims to retrieval the images of the target person from the gallery set across different cameras.
The widely used losses for CNN-based re-id methods include two types: classification loss~\cite{Sun2018Beyond} and metric learning loss~\cite{Hermans2017In}.
Classification loss views re-id in the training stage as a classification task and uses softmax with cross entropy loss to optimize the model.
Sun \etal~\cite{Sun_2020_CVPR} demonstrate that other softmax-based losses, which are widely used in face recognition \eg, AM-Softmax~\cite{Wang_2018_amsoftmax}, CircleLoss~\cite{Sun_2020_CVPR}, are also suitable for re-id.
The most well-known metric learning loss for re-id is triplet loss with hard sample mining~\cite{Hermans2017In}.
For each anchor in a mini-batch, triplet loss with hard sample mining only samples its hardest positive sample and its hardest negative sample to form a triplet to optimize.
To take all samples to participate in optimization, Ristani \etal~\cite{Ristani_2018_CVPR} improve it by proposing an adaptive weighted triplet loss and a new technique for hard-identity mining.

\subsection{Noisy Labels}

Learning with noisy labels has recently drawn much attention as ambiguous and inaccurate labels can exist in most datasets.  Wu \etal~\cite{wu2018light} proposes a semantic bootstrapping method by re-labelling noisy samples by predictions.  Zhong \etal~\cite{zhong2019unequal}  learns discriminative face representation supervised by a noise-resistant loss and copes the long-tail issue by hard identities mining strategy. Hu \etal~\cite{hu2019noise} discovers the distribution of training samples implicitly reflects the probability to be clean and proposes a noise-tolerant end-to-end paradigm by employing the idea of weighting training samples. Co-Mining~\cite{wang2019co} trains twin networks simultaneously, detects noisy labels based on loss values, exchanges high confidence clean faces and re-weight the predicted clean faces. To improve the robustness to label noise of  ArcFace, Sub-center ArcFace~\cite{deng2020sub} relaxes the intra-class constraint by designing K sub-centers for each class  and one training sample only needs to be close to any of the K positive sub-centers instead of the only one positive center. These methods mainly deal with the purity issue which means one class may contain multiple identities. In contrast, our BBS focuses on training the multiple baskets with class overlaps.

\section{Methodology}

\begin{figure*}[!htb]
  \centering
  \includegraphics[clip, trim=0cm 0cm 0cm 0cm, width=0.8\textwidth, height=0.4\textwidth]{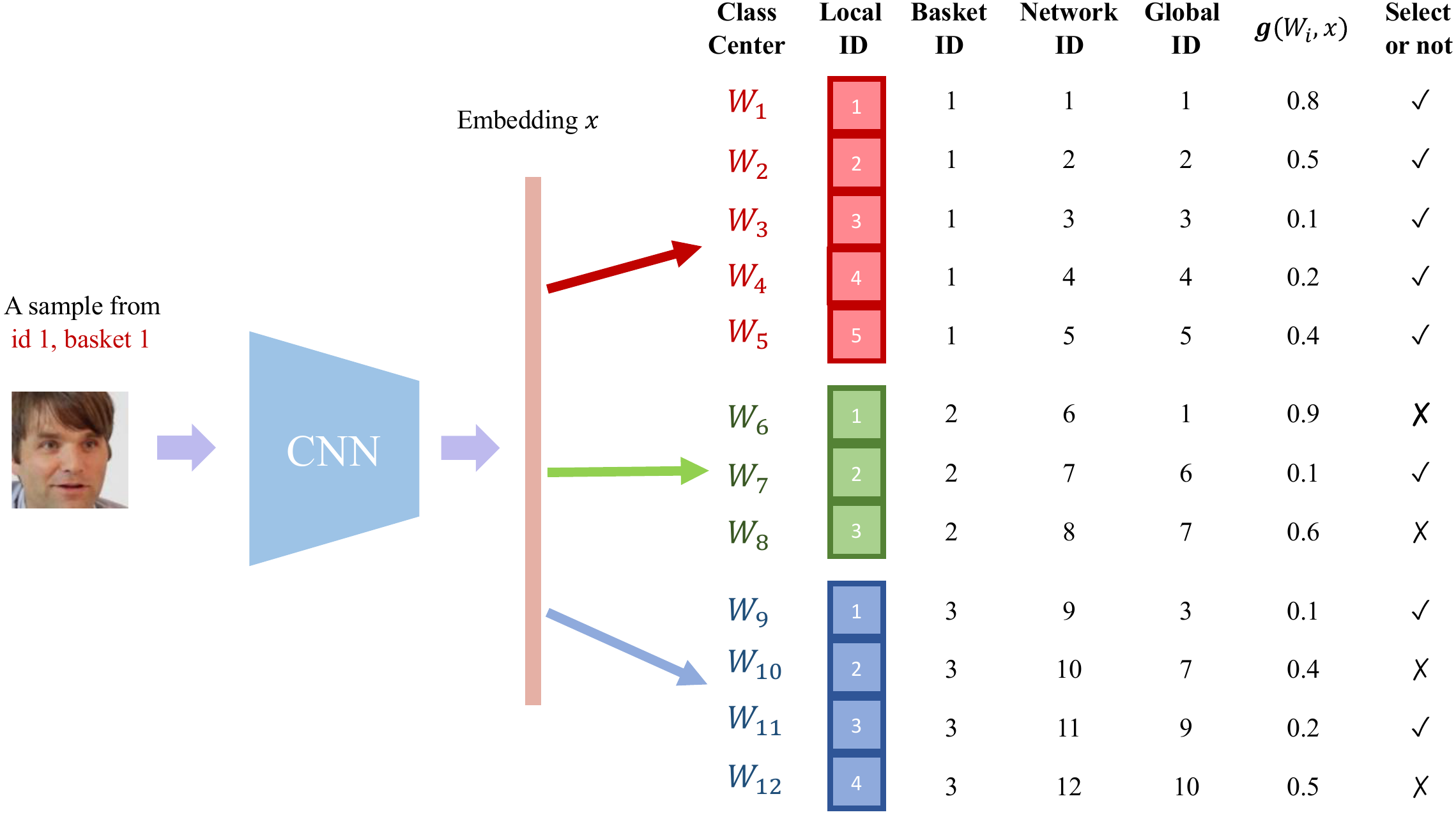}
  \caption{An example for the proposed Basket-based Softmax (BBS) with mining $N_k-2$ negative samples, where $N_k$ is the number of classes in the basket $k$. In this example, there are 3 baskets with 5, 3, 4 classes respectively. Local IDs are concatenated based on basket IDs and form the Network IDs. Global IDs indicate the real class (\textit{e.g.}, images with same global id in face recognition means they are collected from the same person.) In this example, class 2 in basket 2 and classes 1,3 in basket 3 are added to negative classes for current sample.}
  \label{fig:data_example}
\end{figure*}

Our goal is to learn discriminative features from multiple baskets with softmax-based losses, wherein labels inside a basket are clean while class overlaps can exist between baskets. 
To achieve this goal, we propose a novel mining-during-training strategy called Basket-based Softmax (BBS) to effectively train models on multiple baskets in an end-to-end fashion. We first provide a unified perspective for softmax-based losses in  Sec.~\ref{subsec:softmax} and present the details of our BBS in Sec.~\ref{subsec:bbs}.

In real-world applications, dealing with large-scale datasets is another non-negligible problem. To address this, we modified the original BBS to a parallel version in Sec.\ref{subsec:pbbs}, which enables BBS to support \textbf{million} level classes on a single machine.

\subsection{A Unified Perspective of Softmax-based loss} \label{subsec:softmax}

Assuming there are $n$ classes in the training dataset and the feature embedding is of dimension $d$. The softmax-based losses build a fully connected layer with a weight matrix $[W_1, W_2, \cdots, W_n] \in \mathcal{R}^{d\times n}$ and biases $[b_1, b_2, \cdots, b_n]\in \mathcal{R}^{n}$, where each $W_i$ corresponds to the class center $i$. For the sample $i$, denote its class as $y_i$ and the embedding as $x_i$. The softmax-based losses for sample $i$ can be unified into one formula as follows:
\begin{equation}\label{eq:softmax}
  \resizebox{0.4\textwidth}{!}{$
    L = -\log \frac{e^{s(f(W_{y_i}, x_i)+b_{y_i})}}{e^{s(f(W_{y_i}, x_i)+b_{y_i})} + \overset{n}{\underset{
          j=1,
          j\neq y_i
        }{\sum}} e^{s(g(W_{j}, x_i)+b_j)}}
    $}
\end{equation}

In different methods, $f(W, x), g(W, x)$ are defined in various formats as shown in Tab.~\ref{table:method}. The function $g(W, x)$ is used to measure similarity between feature $x$ and class center $W$. Softmax and L-Softmax~\cite{Liu2016} uses the Euclidean distance while NormFace~\cite{Wang2017}, CosFace~\cite{Wang2018} and ArcFace~\cite{Deng2018} uses the cosine similarity. $l_2$-softmax~\cite{Ranjan2017} and SphereFace~\cite{Liu2017} modify the similarities to eliminate the effects of magnitudes of features or class centers.

$f(W,x)$ is either the same as $g(W,x)$, or revised to further increase the intra-class distance as well as decrease inter-class distance (\eg, an additive angular margin is introduced in  ArcFace~\cite{Deng2018}).

\subsection{Basket-based Softmax} \label{subsec:bbs}


Assuming there are $M$ training baskets $S_1, S_2, \cdots, S_M$ and  basket $S_m$ contains $N_m$ classes, $m= 1, 2, \cdots M$.  The local class in basket $m$ starts from $1$ to $N_m$ and we sequentially concatenate labels from all baskets to get the network labels, as shown in Fig.~\ref{fig:data_example}.

Denote $L_0 = 0, L_m = \sum_{i=1}^{m-1}N_i, m=1,2,\cdots M$. Then the total number of network ids is $L_M$. For the sample $x_i$ with basket id $m$ and local id $l$, its network id is $L_{m-1} + l$ in our setting. 
For the sake of expression, we define the following functions:
\begin{equation}
  \label{eq:fg}
  \resizebox{0.3\textwidth}{!}{$
    \begin{array}{ll}
      F(m, l, x_i)&  = e^{s(f(W_{(L_m+l)}, x_i)+b_{(L_m+l)})} \\
      G(m, l, x_i) & = e^{s(g(W_{(L_m+l)}, x_i)+b_{(L_m+l)})}
    \end{array}
    $}
\end{equation}
Then the loss for  basket-based softmax is formulated as follows:
\begin{equation}\label{eq:bbs}
  \resizebox{0.4\textwidth}{!}{$
    L_{bbs}  =  -\log \frac{F(m, l, x_i)}{F(m, l, x_i) + \overset{N_m}{\underset{         \begin{subarray}{c}
            j=1 \\
            j\neq l
          \end{subarray}}{\sum}} G(m,j,x_i) +
      \overset{M}{\underset{
          \begin{subarray}{c}
            k=1 \\
            k\neq m
          \end{subarray}
        }{\sum}} \overset{N_k}{\underset{j=1}{\sum}}  G(k,j,x_i)\cdot p^i_{(L_k+j)}}
    $}
\end{equation}
Here $p^i_{(L_k+j)} \in (0, 1)$  is an indicator whether adding class $j$ in basket $k$  as a negative class for current sample $i$. When $p^i_{(L_k+j)}=0$ for all $k,j$, each basket is assigned with an individual loss and BBS loss degrades to the multi-task approach (\eg, each basket has an individual loss).  In contrast, setting all $p^i_{(L_k+j)}$ to 1 has the same effect of training on concatenated data,  without considering the overlapped classes between baskets.

Function $g(W,x)$ is used as a metric for measuring the similarity between a embedding $x$  and a class center $W$. For each sample $i$ from basket $m$, we select $n_k$ negative classes from basket $k$ ($k = 1,2,\cdots, N_k, k\neq m$) based on the similarities. To be more specific, similarities of current embedding and class centers of basket $k$ are calculated and $n_k$ least similar class centers are picked as negative classes. We summarize the training scheme in Algorithm \ref{alg:bbs}.

\begin{algorithm}
  \caption{Basket-based Softmax}    \label{alg:bbs}
  \textbf{Input}: Training  sets $S_m, m=1,2,\cdots, M$  with  $N_m$ identities and the mining number $n_m$. \\
  \
  \While{NOT end of training}{
    Update training parameters such as learning rate and weight decay, etc. \;
    Take sample $i$ with basket id $m$  and extract its embedding $x_i$\;
    Initialize a $L_M$-length array $p^i$ with 1s\;
    \For {$k = 1, 2, \cdots, M, k\neq m$}{
      \For {$j = 1, 2, \cdots, N_k$}{
        $v_j \leftarrow g(W_{(L_k+l)}, x_i)$\;
      }
      $v = [v_1, v_2, \cdots, v_{N_k}]$ \;
      \For {$j = 1, 2, \cdots, N_k$}{
        \If {$v_{j}$ among  top-$d_k$ of $v$}{
          $p^i_{(L_k+j)} = 0$
        }
      }
    }
    Calculate BBS loss of current data $i$  based on Eq. \eqref{eq:bbs} and update model parameters by back-propagation\;
  }
\end{algorithm}

\subsubsection{Dynamic Negative Class Mining}

A hyperparameter of our BBS is the number of negative class centers $n_k$ from basket $k$. In the ideal situation, $n_k$ can be set to be $N_k-1$ as current sample $i$ belongs to at most one class in that basket. However, this hyperparameter highly relies on the qualities of estimated similarities. If model is not discriminative enough, a smaller $n_k$ should be used. To this end, we design a dynamic mining strategy as shown in Algorithm. \ref{alg:dm}.

We define $d_k = N_k-n_k$ as the number of ignored classes in basket $k$, which indicates that we cannot treat top-$d_k$ similar classes to be negative classes with a high confidence.
We set a minimum ignored number $\tau_k, \tau_k\geq 1$ and an ignored ratio $r_k$. The $r_k$ is dynamically adjusted based on the discriminative ability of the model. Its value is monotonically decreased during our training process.
In the end, we have $d_k =  \max(\tau_k, N_k\cdot r_k)$.

\begin{algorithm}
  \caption{Dynamic mining during training}  \label{alg:dm}
  \textbf{Input}: Minimum ignored number $\tau_k$, the ignored ratio $r_k$ dropped every $t_r$ epochs and $M$ baskets of training  sets $S_1, S_2, \cdots, S_M$.  Training total $T$ epochs.\\
  \For {$t=1,2,\cdots, T$}{
    \For {$k=1,2,\cdots, M$}{
      $r_k = \lceil \frac{T-t}{t_r} \rceil \cdot \frac{t_r}{T} $ \;
      Update $n_k = N_k - \max(\tau_k, N_k\cdot r_k)$ \;
    }
  }
\end{algorithm}

\subsection{Parallel Basket-based Softmax} \label{subsec:pbbs}

For the parallel version of BBS, we first distribute $L_M$ class centers into $G$ GPUs 
with $W_{l_g}, W_{l_g+1}, \cdots, W_{u_g}$ assigned to the $g$-th GPU, where $l_g = g\cdot\lceil\frac{L_M}{G}\rceil$ and $u_g = \min((g+1)*\lceil\frac{L_M}{G}\rceil-1, L_M)$
. For a sample $i$ from basket $m$ and with the network id $y_i$, we have the following parallel BBS:

\begin{equation}\label{eq:pbbs}
  \resizebox{0.42\textwidth}{!}{$
    L_{pbbs}  =  -\log \frac{e^{s(f(W_{y_i}, x_i)+b_{y_i})}}{e^{s(f(W_{y_i}, x_i)+b_{y_i})} +
      \overset{G}{\underset{g=1}{\sum}} \overset{u_g}{\underset{j=l_g}{\sum}}p_j^i\cdot  e^{s(g(W_j, x_i)+b_j)}}
    $}
\end{equation}

Here $p_j^i \in \{0, 1\}$. For classes belongs to basket $m$, then we set $p^i_{j}$ to be 0 if $j=y_i$ and 1 otherwise. The remaining question is how to decide values for classes not in basket $m$.

Because of being distributed, class centers belong to one basket may distribute to multiple GPUs. Consequently, one GPU has no access to all similarities without gathering the scores together. To avoid extra GPU memories consumption, we propose the algorithm~\ref{alg:pbbs} to approximate procedure in lines 6-13 in algorithm~\ref{alg:bbs}. For example, $x_i$ with basket ID $m$, the key idea of the parallel version is to consider each truncated basket as a new one and calculate the number of negative classes based on the new baskets.

\begin{algorithm}
  \caption{Setting $p$ for Parallel BBS}    \label{alg:pbbs}
  Initialize a $L_M$-length array $p^i$ with all elements set to 1 besides
  $p^i_{y_i} \leftarrow 0$\;
  \For {each GPU $g$}{
    \For {$k = 1, 2, \cdots, M, k\neq m$}{
      $l \leftarrow \max(0, L_{m-1} - g_l$) \;
      $u \leftarrow \min(N_G, L_m - g_u)$\;
      \If {$l\geq N_G$ or $u\neq 0$}{
        continue \;
      }
      \For {$j = l, l+1 \cdots, u$}{
        $v_{j} \leftarrow -g(W_{g_l+j}, x_i)$\;
      }
      $v = [v_{l}, v_{l+1}, \cdots, v_{u}]$ \;
      $d^g_k = \max(\tau_k, (u-l+1)\cdot r_k)$ \;
      \For {$j = l, l+1 \cdots, u$}{
        \If {$v_{j}$ is among  top-$d^g_k$ of $v$}{
          $p^i_{l_g+j} = 0$ \;
        }
      }
    }
  }
  Calculate $p_j^i\cdot e^{s(g(W_j, x_i)+b_j)}$  in each GPU and Gather the results \;
  Compute the $L_{pbbs}$ according to Eq.~\ref{eq:pbbs}\;
\end{algorithm}

\section{Experiment}
We evaluate our BBS on two important tasks in the computer vision community: face recognition (section~\ref{subsec:exp_1}) and person re-identification (section~\ref{subsec:exp_2}). For face recognition, we simulate the MBCO problem for experiments on a common-used database called MS1MV2~\cite{Deng2018}. For person re-identification, images from same camera and in same day are gathered to be a basket. As lack of previous works on this problem, we choose two baselines: the first baseline is trained on the concatenated baskets without merging classes, while the second baseline is the multi-task approach with each basket equipped with a softmax loss.  To further verify the capability of handling large-scale data, resource consumptions are studied when modifying BBS to the parallel version (in section~\ref{subsec:exp_3}).

\subsection{Face Recognition}\label{subsec:exp_1}
\begin{algorithm}
  \caption{Split a dataset}    \label{alg:split}
  \textbf{Input}: A dataset $D$. Number of parts $k$ and the probability array $p=[p_1, p_2, \cdots, p_k]$. \\
  \For {each label in $D$}{
    Sample a number $l$ from $[1,2,\cdots, k]$ with probability $p_l$ \;
    Split images of current label into $l$ parts, denote as $s_1, s_2, \cdots, s_l$ \;
    Select $l$ datasets from $D_1, D_2, \cdots, D_k$ and insert image sets $s_1, s_2, \cdots, s_l$ orderly\;
  }
  \textbf{Output}: $D_1, D_2, \cdots, D_k$. \\
\end{algorithm}

To simulate the MBCO problem, we adopt the algorithm~\ref{alg:split} to split a  dataset $D$ into $k$ baskets, where one class appears in $l$ baskets with the probability $p_l$. We modify the state-of-the-art ArcFace~\cite{Deng2018} in face recognition to our BBS framework and conduct experiments by our simulated baskets. We mainly consider two important factors of the MBCO problem: ratio of class overlaps between baskets (in section~\ref{subsec:fr_1}) and number of baskets involved (in section~\ref{subsec:fr_2}).

\subsubsection{Settings}

\noindent \textbf{Datasets.} The MS-Celeb-1M dataset~\cite{guo2016ms} contains about 100k identities with 10 million images. However, it consists of a great many noisy face images. We employ MS1MV2~\cite{Deng2018} (3.8M images, 85k unique identities) as our training dataset. For evaluation, we adopt LFW~\cite{huang2008labeled}, CFP-FP~\cite{sengupta2016frontal}, AgeDB-30~\cite{moschoglou2017agedb}, IJB-B~\cite{whitelam2017iarpa}  and IJB-C~\cite{maze2018iarpa}. All the images are aligned to $112\times 112$ based on 5 facial landmarks, following ArcFace~\cite{Deng2018}.

\medskip
\noindent \textbf{Simulated Datasets.} In section~\ref{subsec:fr_1}, we split the full MS1MV2 dataset into 2 parts with different ratios of class overlaps, from 10\% to 100\%. In section~\ref{subsec:fr_2}, MS1MV2 is split into 10 parts with $p_{i} = 3p_{i+1}$ (\textit{e.g.}, the probability of a class appears $i$ baskets is 3 times of it appears in $i+1$ baskets). In the end, we have $p_i = \frac{2}{3^{10} -1} \cdot 3^{(10-i)}$ and the average ratio of overlaps between two baskets is around 10\%. The average number of classes in one basket is $12810.9$ and experiments are conducted from 2 to 10 baskets to examine the impact of basket numbers.

\medskip
\noindent
\textbf{Training.} We train the models with 8 1080Ti GPUs by stochastic gradient descent (SGD) algorithm. The learning rate is initialized as 0.1 and divided by 10 at 5, 10, 15 epochs, and we finish the training at the 20th epoch.  The weight decay is set to 0.0005 and the momentum is 0.9. We only augment the training samples by random horizontal flip. Because of vast number of identities, the parallel BBS is adopted in the face recognition experiments.

\subsubsection{Ratio of Class Overlaps between Baskets} \label{subsec:fr_1}
\setlength{\tabcolsep}{4pt}
\begin{table*}
  \begin{center}
    \footnotesize{
      \begin{tabular}{c|c|ccccccccc}
        \hline
        Split MS1MV2  & Method &  LFW & CFP-FP & AgeDB  & \multicolumn{3}{c}{IJB-B (TAR@FAR)} & \multicolumn{3}{c}{IJB-C (TAR@FAR)} \\
        \cline{6-11}
        into  &&&&& 1e-5 &  1e-4 & 1e-3 & 1e-5 & 1e-4 & 1e-3 \\
        \hline
        \multirow{3}{*}{2 baskets, overlap 10\%}
                      & {baseline1} & {99.63} & {93.48} & {97.10} & {78.96} & {90.04} & {94.03} & {86.10} & {92.20} & {95.45} \\
                      & {baseline2} & {99.56} & {93.57} & {97.05} & {82.66} & \textbf{90.70} & {94.04} & {88.48} & \textbf{92.83} & {95.51} \\
                      & {BBS} & \textbf{99.68} & \textbf{93.70} & \textbf{97.13} & \textbf{84.23} & {90.68} & \textbf{94.24} & \textbf{89.00} & {92.77} & \textbf{95.66} \\
        \hline
        \multirow{3}{*}{2 baskets, overlap 20\%}
                      & {baseline1} & {99.58} & {93.14} & {96.75} & {83.33} & {90.29} & {94.01} & {88.77} & {92.65} & {97.41} \\
                      & {baseline2} & {99.50} & {93.20} & {97.08} & {82.26} & {90.61} & \textbf{94.07} & {88.20} & {92.62} & {95.47} \\
                      & {BBS} & \textbf{99.67} & \textbf{93.67} & \textbf{97.15} & \textbf{83.80} & \textbf{91.02} & {94.02} & \textbf{89.18} & \textbf{92.94} & \textbf{97.42} \\
        \hline
        \multirow{3}{*}{2 baskets, overlap 30\%}
                      & {baseline1} & {99.60} & {92.64} & {96.81} & \textbf{82.24} & {90.19} & {93.69} & {87.86} & {92.48} & {95.14} \\
                      & {baseline2} & {99.52} & {93.31} & {96.88} & {82.02} & {90.48} & \textbf{94.28} & {88.06} & {92.67} & {95.42} \\
                      & {BBS} & \textbf{99.58} & \textbf{93.47} & \textbf{97.23} & {81.65} & \textbf{90.91} & {94.03} & \textbf{88.07} & \textbf{92.82} & \textbf{95.46} \\
        \hline
        \multirow{3}{*}{2 baskets, overlap 40\%}
                      & {baseline1} & {99.58} & {93.20} & {96.80} & \textbf{81.89} & {89.66} & {93.66} & \textbf{87.77} & {92.17} & {95.13} \\
                      & {baseline2} & {99.52} & {93.10} & {97.01} & {80.24} & {90.05} & {94.09} & {87.32} & {92.29} & {95.47} \\
                      & {BBS} & \textbf{99.62} & \textbf{93.73} & \textbf{96.98} & {80.59} & \textbf{90.60} & \textbf{94.42} & {87.18} & \textbf{92.59} & \textbf{95.65} \\
        \hline
        \multirow{3}{*}{2 baskets, overlap 50\%}
                      & {baseline1} & {99.63} & {92.35} & {96.36} & {80.98} & {88.94} & {93.12} & {87.07} & {91.44} & {94.73} \\
                      & {baseline2} & {99.55} & {93.14} & {97.00} & \textbf{81.77} & {90.34} & {94.13} & {88.11} & {92.47} & {95.50} \\
                      & {BBS} & \textbf{99.68} & \textbf{93.14} & \textbf{97.30} & {81.41} & \textbf{90.56} & \textbf{94.35} & \textbf{88.37} & \textbf{92.80} & \textbf{95.65} \\
        \hline
        \multirow{3}{*}{2 baskets, overlap 60\%}
                      & {baseline1} & {99.53} & {92.18} & {96.08} & {81.11} & {88.49} & {92.67} & {86.42} & {90.78} & {94.32} \\
                      & {baseline2} & {99.65} & {93.07} & {97.03} & {78.74} & {89.95} & {93.97} & {86.76} & {92.37} & {95.45} \\
                      & {BBS} & \textbf{99.70} & \textbf{93.47} & \textbf{97.13} & \textbf{81.28} & \textbf{90.63} & \textbf{94.28} & \textbf{88.33} & \textbf{92.80} & \textbf{95.57} \\
        \hline
        \multirow{3}{*}{2 baskets, overlap 70\%}
                      & {baseline1} & {99.57} & {92.45} & {95.60} & \textbf{79.15} & {87.40} & {92.44} & {84.54} & {90.01} & {94.16} \\
                      & {baseline2} & {99.65} & {92.40} & {96.76} & {71.01} & {87.71} & {93.47} & {81.96} & {90.45} & {94.94} \\
                      & {BBS} & \textbf{99.72} & \textbf{93.43} & \textbf{96.90} & {78.70} & \textbf{89.82} & \textbf{93.98} & \textbf{86.25} & \textbf{92.09} & \textbf{95.36} \\
        \hline
        \multirow{3}{*}{2 baskets, overlap 80\%}
                      & {baseline1} & {99.50} & {91.48} & {94.46} & {73.68} & {85.29} & {91.24} & {81.99} & {88.41} & {93.22} \\
                      & {baseline2} & {99.57} & {91.70} & {96.93} & {75.48} & {85.86} & {92.30} & {81.17} & {88.66} & {93.86} \\
                      & {BBS} & \textbf{99.65} & \textbf{93.21} & \textbf{97.05} & \textbf{81.30} & \textbf{89.75} & \textbf{94.05} & \textbf{86.95} & \textbf{92.19} & \textbf{95.44} \\
        \hline
        \multirow{3}{*}{2 baskets, overlap 90\%}
                      & {baseline1} & {99.50} & {90.40} & {93.81} & {73.38} & {84.32} & {90.72} & {80.09} & {87.62} & {92.92} \\
                      & {baseline2} & {99.32} & {89.22} & {92.30} & {74.11} & {83.60} & {89.14} & {79.87} & {86.80} & {91.51} \\
                      & {BBS} & \textbf{99.67} & \textbf{93.56} & \textbf{96.87} & \textbf{79.75} & \textbf{89.71} & \textbf{94.04} & \textbf{85.84} & \textbf{91.92} & \textbf{95.39} \\
        \hline
        \multirow{3}{*}{2 baskets, overlap 100\%}
                      & {baseline1} & {99.03} & {89.95} & {90.63} & {17.18} & {37.57} & {75.00} & {23.85} & {46.31} & {83.21} \\
                      & {baseline2} & {97.87} & {84.85} & {80.06} & {63.76} & {73.23} & {80.25} & {70.49} & {77.15} & {82.99} \\
                      & {BBS} & \textbf{99.57} & \textbf{93.41} & \textbf{96.92} & \textbf{83.37} & \textbf{90.44} & \textbf{94.25} & \textbf{88.50} & \textbf{92.72} & \textbf{95.64} \\
        \hline
      \end{tabular}
    }
  \end{center}
  \caption{Verification accuracy (\%) on LFW, CFP-FP, AgeDB-30 and IJB-B/IJB-C. Backbone: ResNet18.}      \label{table:fr1}
\end{table*}
\setlength{\tabcolsep}{1.4pt}

Results with different ratios of class overlaps are listed in Tab.~\ref{table:fr1}.  When the ratio equals to $10\%$, BBS surpasses the best results of two baselines by $0.08\%, 0.03\%, 0.03\%$ on LFW, CFP-FP and AgeDB. The numbers are $1.57\%, -0.02\%, 0.20\%$ on IJB-B at TAR@FAR=1e-5, 1e-4, 1e-3 and $0.52\%, -0.05\%, 0.15\%$ on IJB-C at TAR@FAR=1e-5, 1e-4, 1e-3. When the ratio of overlaps increases, BBS has stable high performances while the performances of two baselines drop rapidly. Taking the ratio of $100\%$ as an example, BBS surpasses the best results of two baselines by $0.54\%, 3.56\%, 6.29\%$ on LFW, CFP-FP and AgeDB. The numbers are $20.60\%, 17.21\%, 14.00\%$ on IJB-B at TAR@FAR=1e-5, 1e-4, 1e-3 and $18.01\%, 15.57\%, 12.43\%$ on IJB-C at TAR@FAR=1e-5, 1e-4, 1e-3.

\begin{figure}[htb!]
  \centering
  \includegraphics[clip, trim=0cm 0cm 0cm 0cm, width=0.5\textwidth]{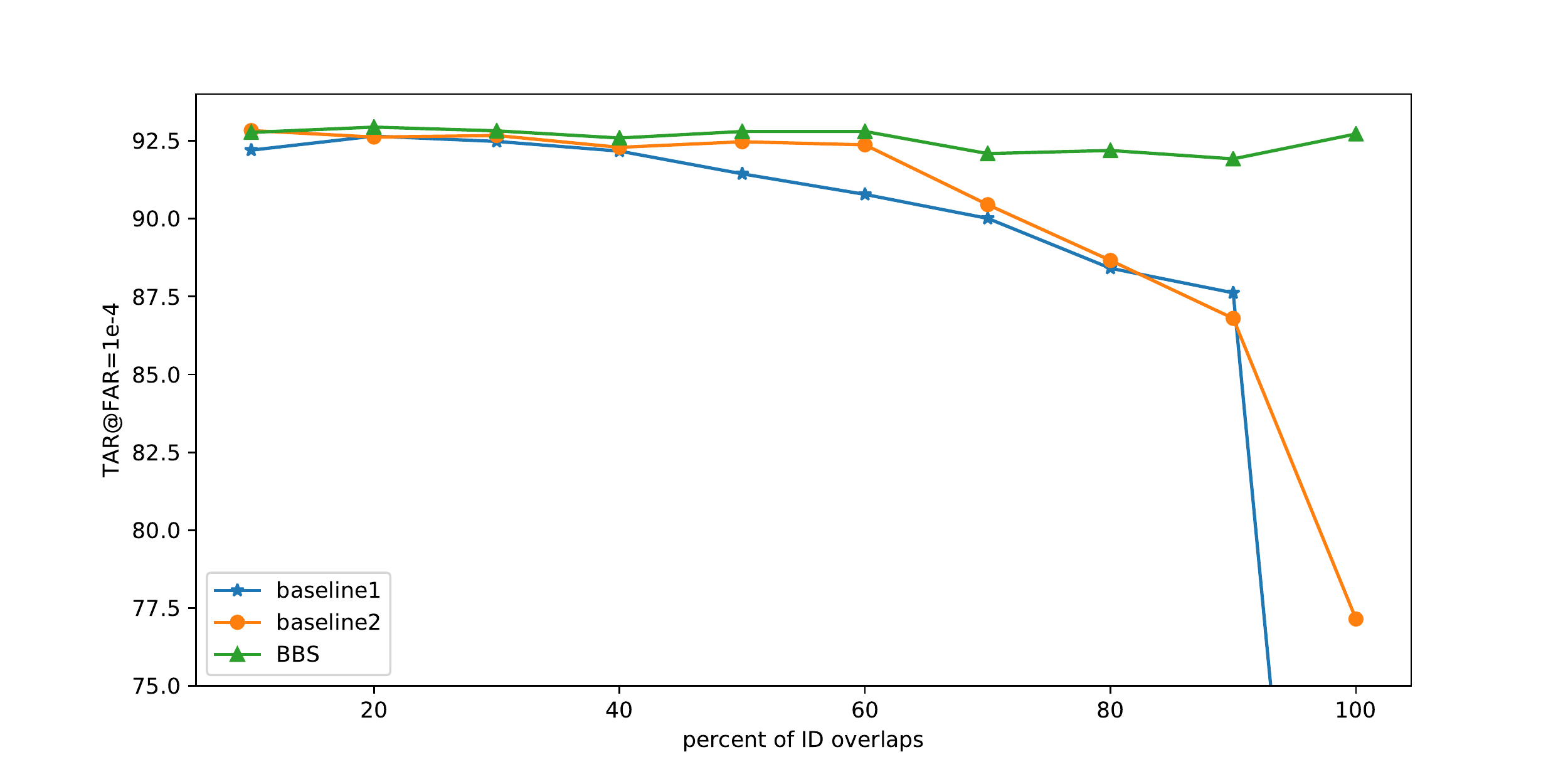}
  \caption{TAR@FAR=1e-4 on IJB-C dataset with different overlaps. With increasing ratios of overlaps, performances of two baselines drop significantly especially with ratio larger than 60\%. In contrast, BBS achieves consistent the best results.}
  \label{fig:percent}
\end{figure}

We further visualize the trend of performances on IJB-C at TAR@FAR=1e-4 in Fig.~\ref{fig:percent}.  With increasing ratio of class overlaps, performance of the baseline1 drops rapidly as the error of labels increases. Labels used by baseline2 are clean as each basket are processed individually. However, with more images separated into different baskets, each class contains less images and that leads to the performance degrade. In contrast, our BBS achieves the stable and best results on TAR@FAR=1e-4 at IJB-C, which shows the superiority of the proposed method compared to the baselines.

\subsubsection{Number of Baskets} \label{subsec:fr_2}

Tab.~\ref{table:fr2} shows the results with different number of baskets. The overall trend is that improvements become more significant with more baskets used. When all baskets used, BBS gets $0.14\%$, $0.33\%$, $0.48\%$ performance boosts compared to the best results from baselines on LFW, CFP-FP and AgeDB. BBS also surpasses the baselines on all TAR criteria by $1.58\%$ on FAR=1e-5, $0.81\%$ on FAR=1e-4 and $0.56\%$ on FAR=1e-3 than the best results from baselines on IJB-B and by $0.56\%$ on FAR=1e-5, $0.34\%$ on FAR=1e-4 and $0.29\%$ on FAR=1e-3 than best results from baselines on IJB-C.  We also train the model with the original MS1MV2 and report the results in the last row of table~\ref{table:fr2}. Even trained with the split datasets, our BBS can still obtain comparable results to the full clean data, which demonstrates the great practical value and efficiency of the proposed method.

\setlength{\tabcolsep}{4pt}
\begin{table*}
  \begin{center}
    \footnotesize{
      \begin{tabular}{l|c|ccccccccc}
        \hline
        Datasets & Method &  LFW & CFP-FP & AgeDB  & \multicolumn{3}{c}{IJB-B (TAR@FAR)} & \multicolumn{3}{c}{IJB-C (TAR@FAR)} \\
        \cline{6-11}
                 &&&&         & 1e-5 &  1e-4 & 1e-3 & 1e-5 & 1e-4 & 1e-3 \\
        \hline
        \multirow{3}{*}{2 baskets}
                 & {baseline1} & {99.45} & {88.80} & {95.30} & \textbf{78.46 } & {86.32} & {91.57} & \textbf{83.78} & {89.01} & {93.49} \\
                 & {baseline2} & {99.38} & {89.00} & {95.45} & {74.26} & {86.38} & {91.71} & {83.02} & {89.11} & {93.51} \\
                 & {BBS} & \textbf{99.43} & \textbf{89.70} & \textbf{95.51} & {77.16} & \textbf{86.49} & \textbf{91.88} & {83.68} & \textbf{89.14} & \textbf{93.57} \\
        \hline
        \multirow{3}{*}{4 baskets}
                 & {baseline1} & {99.38} & {90.46} & {96.11} & \textbf{81.64} & {88.50} & {92.94} & \textbf{86.97} & {91.22} & {94.49} \\
                 & {baseline2} & {99.55} & {91.21} & {96.08} & {79.70} & {88.63} & {93.01} & {86.82} & {91.23} & {94.67} \\
                 & {BBS} & \textbf{99.60} & \textbf{91.63} & \textbf{96.37} & {80.84} & \textbf{88.79} & \textbf{93.17} & {86.50} & \textbf{91.42} & \textbf{94.71} \\
        \hline
        \multirow{3}{*}{6 baskets}
                 & {baseline1} & {99.47} & {91.50} & {96.30} & {82.06} & {88.97} & {93.05} & {86.89} & {91.39} & {94.56} \\
                 & {baseline2} & {99.45} & {92.56} & {96.42} & {79.32} & \textbf{89.49} & {93.37}&  {86.38} & \textbf{91.83} & {94.98} \\
                 & {BBS} & \textbf{99.60} & \textbf{92.57} & \textbf{96.72} & \textbf{82.42} & {89.31} & \textbf{93.67} & \textbf{86.95} & {91.77} & \textbf{95.12} \\
        \hline
        \multirow{3}{*}{8 baskets}
                 & {baseline1} & {99.38} & {92.63} & {96.52} & {80.80} & {88.72} & {93.11} & {86.81} & {91.20 } & {94.71 } \\
                 & {baseline2} & {99.63} & {93.20} & {96.50} & {80.05} & {89.80} & {93.65} & {86.78} & {91.93} & {95.12} \\
                 & {BBS} & \textbf{99.72} & \textbf{93.51} & \textbf{96.67} & \textbf{81.19} & \textbf{90.01} & \textbf{93.81} & \textbf{87.60} & \textbf{92.16} & \textbf{95.23} \\
        \hline
        \multirow{3}{*}{10 baskets}
                 & {baseline1} & {99.43} & {92.41} & {96.52} & {80.45} & {89.61} & {93.27} & {87.35} & {91.79} & {94.87} \\
                 & {baseline2} & {99.53} & {93.51} & {96.65} & {78.95} & {89.90} & {93.85} & {86.74} & {92.31} & {95.34} \\
                 & {BBS} & \textbf{99.67} & \textbf{93.84} & \textbf{97.03} & \textbf{82.03} &\textbf{90.71} &\textbf{94.29} & \textbf{87.91} & \textbf{92.65} & \textbf{95.63} \\
        \hline
        MS1MV2    & {ArcFace } & {99.62} & {93.20} & {96.82} & \textbf{83.83} & \textbf{91.13} & {94.22} & \textbf{89.44} & \textbf{93.20} & \textbf{95.57} \\
        \hline
      \end{tabular}
    }
  \end{center}
  \caption{Verification accuracy (\%) on LFW, CFP-FP, AgeDB-30 and IJB-B/IJB-C. Backbone: ResNet18.}      \label{table:fr2}
\end{table*}
\setlength{\tabcolsep}{1.4pt}

\subsection{Person Re-identification}\label{subsec:exp_2}

In this section, we verify the generalization of our proposed BBS on person re-identification. Specifically, three losses, including Softmax with cross entropy loss, CosFace~\cite{Wang2018} loss, and ArcFace~\cite{Deng2018} loss, are used as the training loss respectively to verify the generalization of the proposed BBS.

\subsubsection{Settings}
\noindent
{\bf Dataset}
Market-1501 dataset~\cite{Zheng2015Scalable} contains 1501 identities and 32217 images captured by 6 cameras.
Following \cite{Zheng2015Scalable}, 751 identities are reserved for training and the remaining 750 identities are used for testing.
All these data are captured from six time periods.
To verify the effectiveness of the proposed BBS, we divide the training set of Market-1501 into six baskets according to the capture time and name it Market-Basket dataset.
In the testing stage, we use the original testing set of Market-1501 as the testing set of Market-Basket for evaluation.

\medskip
\noindent
{\bf Implementation Details.}
Following \cite{Hou2019Interaction}, We use ResNet-50~\cite{He2016Deep} as the backbone.
The model is trained for 60 epochs in total by Adam~\cite{Kingma2014Adam} optimizer.
The learning rate is initialized as $3.5\times10^{-4}$ and multiplied by 0.1 after every 20 epochs.
The batch size is set to 64, and each batch consists of 16 persons and 4 images for each person.
All these images are resized to $256\times128$ pixels.
We also use horizontal flip, random crop, and random erase~\cite{zhong2020random} for data augmentation.
As for CosFace loss and ArcFace loss, the scaling factor $s$ is set to 16 and the margin $m$ is set to 0.1 by grid search.
For BBS, $(\tau_k, t_r)$ are set to (2,2), respectively.

\medskip
\noindent
{\bf Evaluation Protocol.} Cumulative Matching Characteristics (CMC) and mean Average Precision (mAP) are used as the evaluation metrics.

\setlength{\tabcolsep}{5pt}
\begin{table}
  \begin{center}
    \small{
      \begin{tabular}{l|c|ccc}
        \hline
        Loss & Method &  top-1 & top-5 & mAP \\
        \hline
        \multirow{3}{*}{Softmax}
             & {baseline1} & {85.4} & {94.2} & {65.9} \\
             & {baseline2} & {78.4} & {89.9} & {53.9} \\
             & {BBS} & \textbf{86.0} & \textbf{94.7} & \textbf{67.8} \\
        \hline
        \multirow{3}{*}{CosFace~\cite{Wang2018}}
             & {baseline1} & {89.0} & {85.9} & {73.6} \\
             & {baseline2} & {85.3} & {93.2} & {64.0} \\
             & {BBS} & \textbf{90.8} & \textbf{96.4} & \textbf{76.6} \\
        \hline
        \multirow{3}{*}{Arcface~\cite{Deng2018}}
             & {baseline1} & {88.5} & {96.0} & {72.7} \\
             & {baseline2} & {84.9} & {91.8} & {63.4} \\
             & {BBS} & \textbf{89.7} & \textbf{96.3} & \textbf{75.9} \\
        \hline
      \end{tabular}
    }
  \end{center}
  \caption{The results with different losses on Market-Basket.}
  \label{table:reid}
\end{table}

\subsubsection{Generalization for Different Losses}
As shown in Table~\ref{table:reid}, BBS surpasses baseline1 and baseline2 significantly and consistently for all these three losses.
Specifically, BBS increases about 2\% top-1 and 3\% mAP over baseline1 for CosFace, which does not mine any cross-basket negative class.
This comparison demonstrates the effectiveness of BBS for dynamic negative class mining and the generalization of BBS for different softmax-based losses.

\subsection{Parallel  Acceleration}\label{subsec:exp_3}

\begin{figure}[htb!]
  \centering
  \subfloat[GPU Memory\label{fig:parallel_gpu}]{\includegraphics[width=0.38\textwidth]{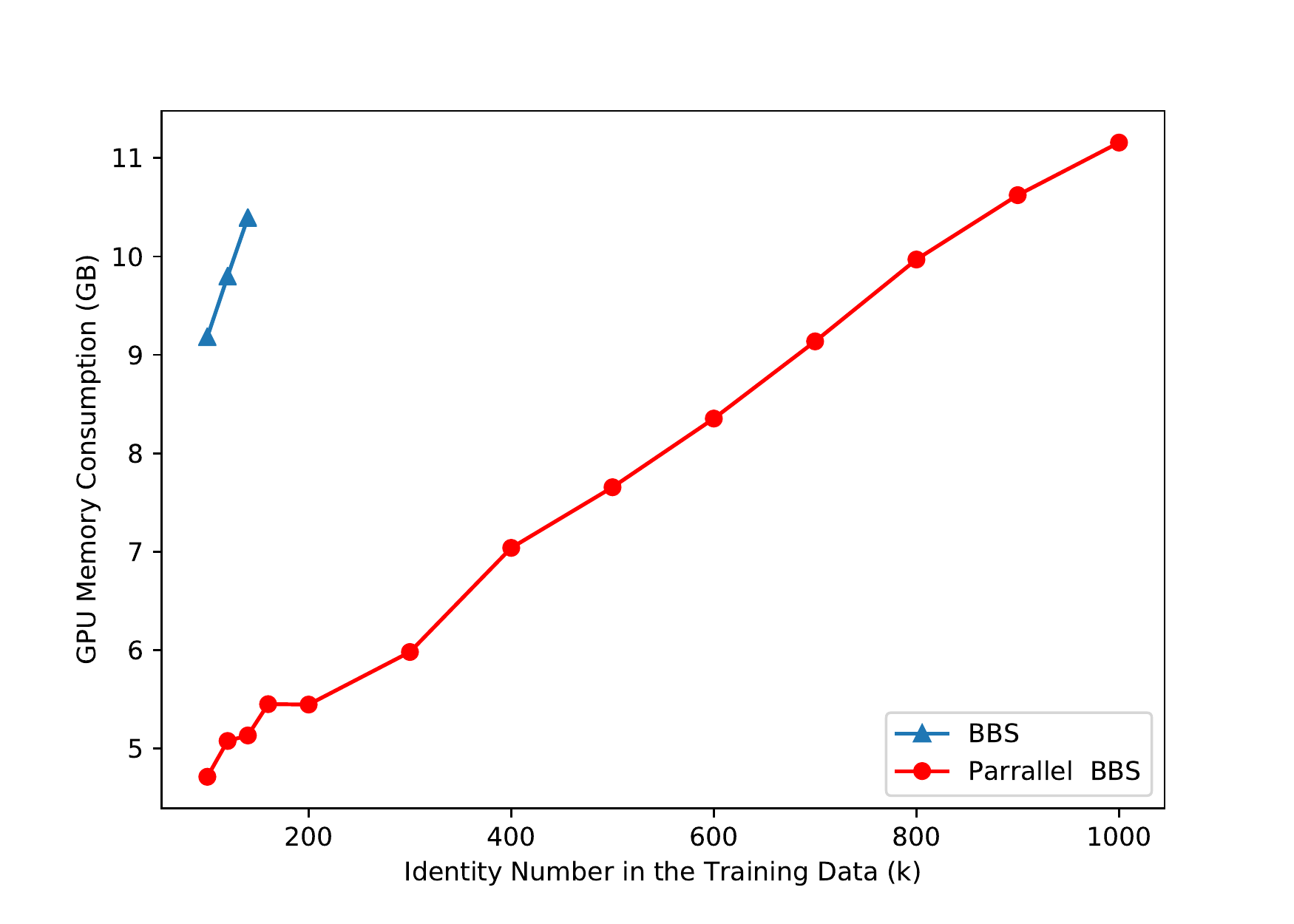}} \\
  \subfloat[Training Speed\label{fig:parallel_to}]{\includegraphics[width=0.38\textwidth]{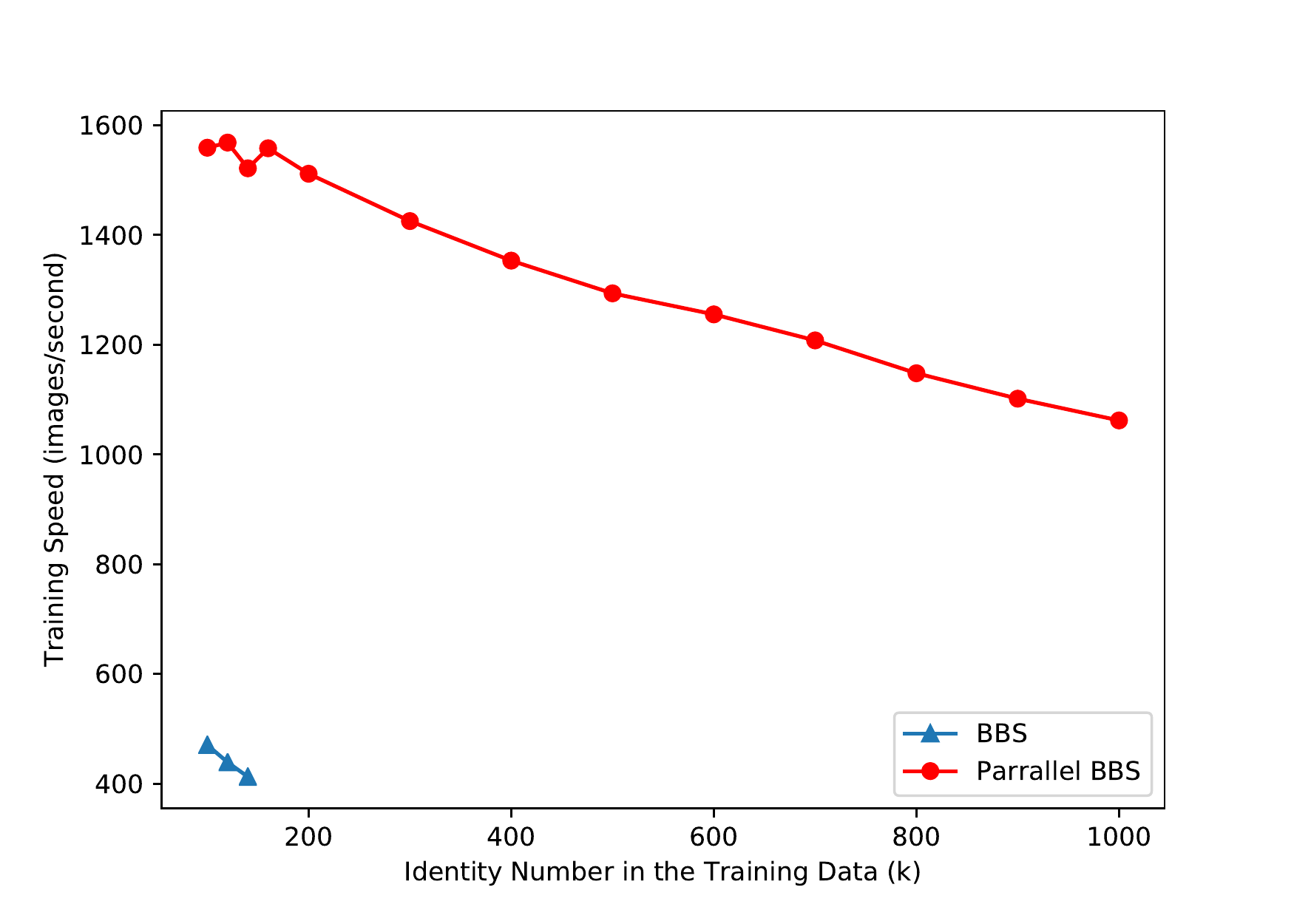}}
  \caption{Parallel acceleration. Setting: ResNet 18, batch size 512, feature dimension 512, 8 GPU 1080Ti (12GB).}
  \label{fig:parallel}
\end{figure}

We simulate the trainings on different number of classes on 8 GPUs (1080Ti with 12GB memory) to test the training speed of the BBS and parallel BBS and visualize the results in Fig.~\ref{fig:parallel}. The original BBS can train less than 200k classes and the throughputs are smaller than 500 images per second. In contrast, the parallel BBS is able to handle datasets with up to 1M classes and the throughputs can be 3 times those of the original BBS. The throughputs remain more than 1000 images per second even trained with 1M classes.

\section{Conclusion}

In this paper, we raise the importance of the Multiple Baskets with Class Overlaps (MBCO) problem which is usually ignored by the academic community but can occur frequently in real-world applications, and propose an end-to-end mining-during-training framework called Basket-based Softmax (BBS) to enable the training on multiple baskets.  Extensive experiments are conducted on face recognition and person re-identification, and have verified the superiority, efficiency as well as the generalization of our proposed method.

{\small
  \bibliographystyle{ieee_fullname}
  \bibliography{egbib}
}

\end{document}